\documentclass[runningheads]{llncs}
\usepackage{microtype}
\usepackage{lipsum}
\usepackage[T1]{fontenc}
\usepackage{graphicx}
\usepackage{amsmath}  
\usepackage{algpseudocode}  
\usepackage{algorithm}  
\usepackage{algorithmicx}  
\usepackage{url}
\usepackage[numbers]{natbib}
\begin{document}
\title{MultiClear: Multimodal Soft Exoskeleton Glove for Transparent Object Grasping Assistance}
%
%
\author{Chen Hu\inst{1}\orcidID{0009-0005-9471-1632} \and
Timothy Neate\inst{1}\orcidID{0000-0002-1387-8168} \and
Shan Luo\inst{1}\orcidID{0000-0003-4760-0372} \and
Letizia Gionfrida\inst{1,2}\orcidID{0000-0002-0992-6526}}
\authorrunning{C. Hu et al.}
%
\institute{King's College London, London, WC2R 2LS, UK 
\email{\{tyrone.hu, timothy.neate, shan.luo, letizia.gionfrida\}@kcl.ac.uk}\\
 \and
Harvard University, Cambridge, MA, USA\\
\email{gionfrida@seas.harvard.edu}}
\maketitle              
\markboth{MultiClear}{MultiClear}

\begin{abstract}
Grasping is a fundamental skill for interacting with the environment. However, this ability can be difficult for some (e.g. due to disability). Wearable robotic solutions can enhance or restore hand function, and recent advances have leveraged computer vision to improve grasping capabilities. However, grasping transparent objects remains challenging due to their poor visual contrast and ambiguous depth cues. Furthermore, while multimodal control strategies incorporating tactile and auditory feedback have been explored to grasp transparent objects, the integration of vision with these modalities remains underdeveloped. This paper introduces \emph{MultiClear}, a multimodal framework designed to enhance grasping assistance in a wearable soft exoskeleton glove for transparent objects by fusing RGB data, depth data, and auditory signal. The exoskeleton glove integrates a tendon-driven actuator with an RGB-D camera and a built-in microphone. To achieve precise and adaptive control, a hierarchical control architecture is proposed. For the proposed hierarchical control architecture a high-level control layer provides contextual awareness, a mid-level control layer processes multimodal sensory inputs, and a low-level control executes PID motor control for fine-tuned grasping adjustments. The challenge of transparent object segmentation was managed by introducing a vision foundation model for zero-shot segmentation. The proposed system achieves a Grasping Ability Score of 70.37 ± 3.96\%, demonstrating its effectiveness in transparent object manipulation. 

\end{abstract}
\section{Introduction}

Grasping is a fundamental skill required for interacting with objects in daily life. However, individuals with hand impairments \cite{raghavan2007nature}, such as those resulting from stroke, spinal cord injury, or neuromuscular disorders, often experience significant difficulties in generating stable and controlled grasps \cite{raghavan2007nature, Silver2021}. These impairments limit autonomy and quality of life, necessitating assistive solutions to restore hand function. Wearable robots, particularly exogloves (exogloves), have been developed to augment grasping ability by applying external torque to facilitate finger flexion and extension towards a more stable grasp \cite{gionfrida2024wearable, kim2019eyes}. Compared to rigid ones, soft exogloves constructed from flexible materials offer enhanced comfort and adaptability, making them suitable for prolonged use in daily activities \cite{coyle2018bio, du2021review}.

Recent advancements in wearable robotics have explored the integration of vision to enhance adaptability and environmental awareness in robotic assistance \cite{rho2024multiple, missiroli2023integrating}. Visual cues can provide contextual information that enables intent recognition and real-time grasp modulation, reducing reliance on direct user input. However, existing vision-assisted exogloves primarily focus on opaque objects, struggling with transparent objects due to low visual contrast and ambiguous depth cues. This limitation impacts their applicability in real-world scenarios, where transparent objects such as glassware are common.

\begin{figure*} [t]
  \includegraphics[width=\textwidth]{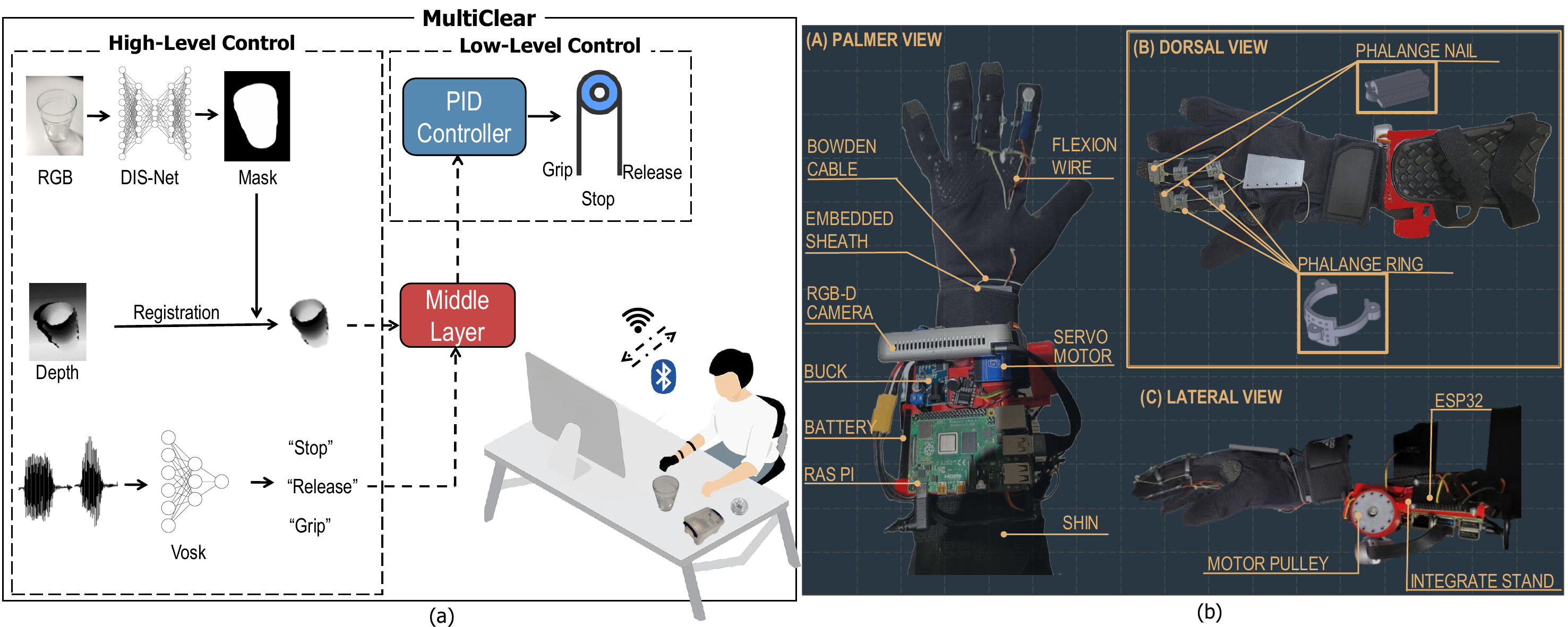}
  \caption{(a) \textbf{Overview of \emph{MultiClear}: } The proposed multimodal framework captures RGB, depth and auditory inputs through sensors mounted on the exoglove. The high-level control module processes the RGB frames using DIS-Net \cite{qin2022highly} for forward inference, generating a mask of the target object. The mask is aligned with the current depth frame to extract depth information, which is then passed to the middle layer. Simultaneously, voice input is processed by the Vosk ASR model \cite{vosk2020}, which converts predefined prompt words into control commands. These commands are fed into the middle layer. By integrating the data from all three modalities, the middle layer makes decisions and sends target commands to the low-level PID controller for motor rotation, executing grasp or release actions. (b) \textbf{Hardware setup: } The soft exoglove consists of three main components: an actuator, a tendon-driven glove, and sensing equipment. 3D-printed custom components were designed for routing the wires, connected to the motor to trigger grasp and release actions. After gathering data from the three modalities, client-server interaction is facilitated via Wi-Fi and Bluetooth modules in the microcontrollers.}
  \label{fig:overview}
  \vspace{-0.6cm}
\end{figure*}

Depth completion and estimation have been studied \cite{sajjan2020clear,tang2021depthgrasp,sun2024diffusion}. ClearGrasp \cite{sajjan2020clear} predicts the surface normal and the transparent boundary to solve the depth estimation. Duisterhof et al. \cite{duisterhof2024residual} use NeRF to recover depth to grasp transparent objects. Approaches based on generative models, such as GANs \cite{tang2021depthgrasp} and Diffusion models \cite{sun2024diffusion} were proposed to generate sparse parts of the depth map. Although significant progress has been made for depth completion, these methods remain computationally expensive, preventing real-time deployment, and they are primarily designed for robotic manipulation. In robotic grasping tasks, complete depth information of transparent objects is essential for computing critical grasp-related parameters, such as 6D poses of both the target object and the robotic arm, as well as grasp planning. However, in grasp assistance, intent originates from the user rather than an autonomous system. This implies that the intent detection algorithm only needs to leverage a limited amount of environmental information to timely trigger user intent, rather than performing comprehensive depth reconstruction.

Futhermore, while multimodal control strategies incorporating tactile and auditory feedback have been investigated to improve intent detection and grasping stability \cite{rho2021learning, tran2022flexotendon}, the integration of vision with these modalities remains underexplored. Current control paradigms often rely on single-modality inputs, such as surface electromyography signals \cite{sierotowicz2022emg} or force-sensing resistors \cite{rho2021learning}, which provide limited contextual awareness and require explicit user effort to initiate grasping actions. 

To address these challenges, \emph{MultiClear}, a multimodal framework designed to enhance grasping accuracy for transparent objects by integrating RGB, depth, and auditory modalities was introduced. The system consists of a tendon-driven soft exoglove equipped with an RGB-D camera and a built-in microphone. A hierarchical control architecture that operates across three layers to achieve precise and adaptive control was proposed to: 1) High-level control provides contextual awareness by processing multimodal sensory inputs. 2) Mid-level control fuses data from RGB, depth, and auditory sources for intent recognition. 3) Low-level control executes fine-tuned grasping adjustments using proportional-integral-derivative (PID) motor control. The contributions of this work are:

\begin{enumerate}
\vspace{-0.3cm}
\item To introduce a vision foundation model for zero-shot segmentation to detect and segment the boundaries of transparent objects. Aligning these boundaries with the depth map facilitates grasp point extraction and intent triggering, enhancing grasping accuracy and reliability.
\item To introduce a multimodal hierarchical control framework that integrates RGB, depth, and auditory, deployed on an in-house developed soft exoglove. This includes a high-level control layer for contextual awareness, a mid-level control layer for multimodal sensor fusion, and a low-level control layer for precise motor execution using PID control.

\end{enumerate}

\section{Related Work}

\subsection{Controllers for Soft Exoglove}
Recent advancements in upper-limb wearable robots have introduced diverse control strategies to enhance user interaction, adaptability, and functionality \cite{triwiyanto2021review, du2021review}. Traditional control methods, including joystick-based approaches, have been widely implemented due to their simplicity and reliability \cite{sierotowicz2022emg}. However, these systems often exhibit limitations in dexterity, rendering them unsuitable for executing complex or highly nuanced movements \cite{alicea2021soft}. Force-feedback controllers have demonstrated efficacy in telemanipulation applications by improving user control based on haptic feedback\cite{baselli2024tendon}. Nevertheless, such systems rely on physical contact with objects before actuation, which restricts their adaptability and hampers their ability to provide task-specific assistance in diverse scenarios \cite{baselli2024tendon}.

Emerging neuromuscular interface-based control strategies have garnered substantial attention for their capacity to decode user intent from muscle activity, enabling proportional and adaptive assistance \cite{sierotowicz2022emg, lotti2020intention}. While these approaches hold significant promise for facilitating intuitive user interactions, challenges remain in optimising their integration to ensure seamless and robust performance in dynamic, real-world environments. 
To address these challenges, recent efforts have focused on equipping wearable robotic systems with the ability to perceive and adapt to the surrounding context \cite{gionfrida2024wearable}. By leveraging visual data, these systems can dynamically adjust assistance levels and anticipate user intentions, leading to improved adaptability and enhanced planning capabilities. Compared to other modalities, vision-based perception has been shown to mitigate inter-user and inter-trial variability while improving responsiveness and contextual awareness \cite{tricomi2023environment, kim2019eyes}. These advancements highlight the growing potential of context-aware systems to redefine the interaction paradigms of wearable robots, ensuring greater user satisfaction and task-specific adaptability.

\subsection{Multimodal Fusion}
Humans perceive the world through multiple sensory modalities, such as vision, and hearing, allowing them to form a wide understanding of their environment\cite{zhang2024multimodal}. With advancements in sensor technology, it has become increasingly feasible to collect and analyze diverse forms of data for downstream tasks. However, integrating these heterogeneous data streams effectively remains a challenge, particularly in real-world environments where such data may be noisy, imbalanced, or even corrupted. For traditional assistive devices, empirical and theoretical studies have demonstrated that multimodal fusion approaches may fail under such conditions due to their inability to effectively manage the complexity of heterogeneous sensory inputs and dynamically adapt to variations in data quality \cite{peng2022balanced, xu2022different, huang2021learning}. This failure often arises because conventional approaches treat multimodal data at a single decision level, making it difficult to prioritize or reweight sensory information based on contextual relevance or reliability. These limitations highlight the need for a hierarchical control architecture that can decompose the decision-making process across multiple levels, enabling more robust and adaptive responses by contextualizing high-level information, fusing multimodal data efficiently, and ensuring precise low-level motor execution. To address these challenges, the proposed multimodal framework introduces a hierarchical control architecture that integrates RGB, depth, and auditory data, to enhance contextual awareness and ensure robust, adaptive control during grasping tasks. By incorporating a high-level control layer for contextual understanding, a mid-level control layer for multimodal sensor fusion, and a low-level control layer for precise motor execution through PID control, the proposed framework aims to deliver reliable assistive control strategies, even in complex and uncertain scenarios.

\section{Hardware Design and Control}

\subsection{Hardware Design}
The implemented soft exoglove, based on an existing design \cite{rho2021learning}, consists of three primary components (Fig. \ref{fig:overview} (b)): 1) an embedded actuator utilizing a Tendon-Sheath Mechanism (TSM), 2) a custom glove that transfers force to the finger joints, and 3) a sensing module for interaction feedback. The actuation system, powered by a LiPo battery (Crazepony, 1400mAh, 11.1V, 64.1g, Shenzhen, China), weighs 0.5 kg and is mounted on a forearm-worn shin guard (Super Comfortable Shin Pad, Northdeer, China). A flat servo motor (Digital Servo, 4.8V, 24 kgcm, CHICIRIS, China) drives a 30mm diameter pulley, around which the actuation cable is wound.


The system employs a microcontroller (Raspberry Pi 4B, 1.5GHz, 4GB RAM, Broadcom, UK) for acquiring RGB-D data with a main control chip (ESP32, 2.4GHz, 4GB, Bluetooth, Shanghai, China). A voice module (MARKELL, China) converts analogue inputs into digital signals for further processing.

The TSM connects the actuation system to the glove and is designed to remain fixed, reducing the impact of dynamic sheath bending angles. To maintain a tight, straight connection between the actuation system and the palm-supporting pieces, the sheaths are slightly tensioned, which minimizes sheath bending during operation. This approach reduces unwanted contact between the tendon wires and the sheath, effectively stabilizing fingertip force against dynamic changes in the sheath’s bending angle.

The glove itself features 3D-printed rings and nails installed on the metacarpophalangeal (MCP), proximal interphalangeal (PIP), and distal interphalangeal (DIP) joints, simulating the passage of human flexion tendons through these joints. This design enables the glove to perform grasping and releasing motions. Flexion tendons are routed through the palm, while extension tendons are positioned on the back of the hand. These tendons are pulled and released antagonistically, driven by the motor’s clockwise and counterclockwise revolutions, allowing the index and middle fingers to curl inward and release, respectively. 



A RealSense D415 camera (Intel RealSense, California, USA) is mounted beneath the wrist to capture environmental data. The camera, attached to the shin pad via a miniature gimbal, sends color and depth images through the Raspberry Pi’s Wi-Fi module to a server (Dell, Precision 7680, US) for multimodal inference. A microphone (Uxcell, China) is integrated into the actuation system to capture system inputs. 


\subsection{Heirarchical Control Strategy}
With the integration of an RGB-D camera and a microphone, the actuation system supports multimodal inputs (Fig. \ref{fig:overview} (a)), including RGB frames, depth frames, and voice signals. The high-level control module within the multimodal framework processes visual and auditory data concurrently. Once the RGB-D camera mounted on the exoglove receives the data stream, the system transmits it via Wi-Fi to the server, where the data is processed frame by frame. To optimize efficiency, one frame is selected for inference every three frames. The selected RGB frame is processed by the dichotomous image segmentation network (DIS-Net), generating a mask of the target object. This mask is then aligned with the current depth frame to extract depth information of the object's boundary, which is forwarded to the middle layer for further processing. At the edges of transparent objects, abrupt normal variations and thickness changes create high-contrast features, resulting in depth sparsity rarely occurring at the edges, while it is predominantly observed in the interior regions of transparent objects. Consequently, the RealSense camera can still capture depth information at the object’s edges.

\begin{algorithm}[t]
\scriptsize
\caption{middle layer Control Algorithm}
\begin{algorithmic}[1]
    \State \textbf{Initialize} stack for storing commands
    \State \textbf{Initialize} torqueDeadZone $\gets$ predefined value
    \State \textbf{Initialize} pendingGripCommand $\gets$ False

    \While{True}
        \State \# Check if "grip" command is received
        \If{receivedCommand == "grip"}
            \State Push "grip" to stack
            \State pendingGripCommand $\gets$ True
            \State \# Process depth information if a "grip" command is active
            \State currentDepth $\gets$ Get current depth from Depth Frame
            \State dist $\gets$ Calculate distance between graspingPoint and camera
            \If{dist $\leq$ distThreshold}
                \State Wait for 2 seconds
                \State Send "grip" command via Bluetooth to low-level PID Controller
                \State pendingGripCommand $\gets$ False
            \EndIf
        \EndIf

        \State \# Check if "release" command is received
        \If{receivedCommand == "release"}
            \State Send "release" command to PID Controller
            \If{"grip" exists in stack}
                \State Remove "grip" from stack
            \EndIf
        \EndIf

        \State \# Check if "stop" command is received
        \If{receivedCommand == "stop"}
            \State Send "stop" command to PID Controller
            \State pendingGripCommand $\gets$ False
        \EndIf

        \State \# Discard depth and torque values unless a "grip" command is active
        \If{pendingGripCommand == False}
            \State Discard current depth and torque values
        \EndIf
    \EndWhile
\end{algorithmic}
\end{algorithm}

Voice signals, after analog-to-digital conversion, are input into the Vosk ASR model \cite{vosk2020} for real-time transcription. Vosk leverages a Time-Delay Neural Network (TDNN) \cite{waibel2013phoneme} and Long Short-Term Memory (LSTM) \cite{hochreiter1997long} architecture to map the voice input into text. The three command prompts defined as --  ``\emph{grip}'',  ``\emph{release}'', and  ``\emph{stop}'' -- were sent to the middle layer for interpretation. 

The middle layer discards depth information and torque values unless a  ``\emph{grip}'' command is received. Upon receiving the  ``\emph{grip}'' command, it starts processing the object’s depth information and pushes the command onto a stack. According to depth information at the object's boundaries, the object centroid using Principal Component Analysis (PCA) \cite{Pearson1901PCA} was computed and defined as the grasp point. The grasp point is continuously detected in real-time to update the distance between the exoskeleton glove and the object, thereby adapting to and autonomously triggering the user's grasp intent. The middle layer delays for two seconds before sending the ``grip'' command via Bluetooth to the low-level velocity PID Controller. Simultaneously, it checks if the real-time torque has reached a predefined dead zone, in which case a  ``\emph{Maintain}'' command is issued to the PID Controller. Upon receiving a  ``\emph{release}'' command, the middle layer forwards it to the PID controller and removes any pending  ``\emph{grip}'' command from the stack. If a  ``\emph{stop}'' command is received, it is immediately transmitted to the PID Controller.

The low-level PID Controller, when receiving the  ``\emph{grip}'' command, initiates forward motor rotation, applying a constant torque of 24 kgcm ($\approx$ 2.35 Nm) for fine-tuned grasping adjustment when the power system provided 4.8V voltage. Based on the benchmarking comparison, this precise torque aided users in grasping objects. According to \cite{polygerinos2015soft}, the required force to grasp objects during ADLs does not exceed 15 N, and the pinch forces required to execute most of the daily life tasks are lower than 10.5 N \cite{smaby2004identification}. When the  ``\emph{Maintain}'' command is active, the system ignores all other voice inputs except ``release,'' preventing environmental noise from interfering with PID control and causing grasp failure or slippage. Upon receiving the  ``\emph{release}'' command, the controller reverses motor rotation to release the object. The  ``\emph{stop}'' command halts motor operation immediately.

\vspace{-0.3cm}
\section{Research Protocol}
\vspace{-0.3cm}
To evaluate the performance of \emph{MultiClear} with users, a case study involving six participants with approval (\textit{MRPP-23/24-40750}) from the University and the College Research Ethics Committee (CREC) at King’s College London,  to test the implemented exoglove was conducted. All participants were instructed to keep their hands relaxed and refrain from applying any force to the six transparent objects during the Grasping Ability Score (GAS) test. Additionally, feedback from the users were gathered and summarized. To demonstrate the effectiveness of the vision foundation model for grasping assistance, the intermediate results from \emph{MultiClear} was also visualized.

\begin{figure*}[t]
\centering
\includegraphics[width=\textwidth]{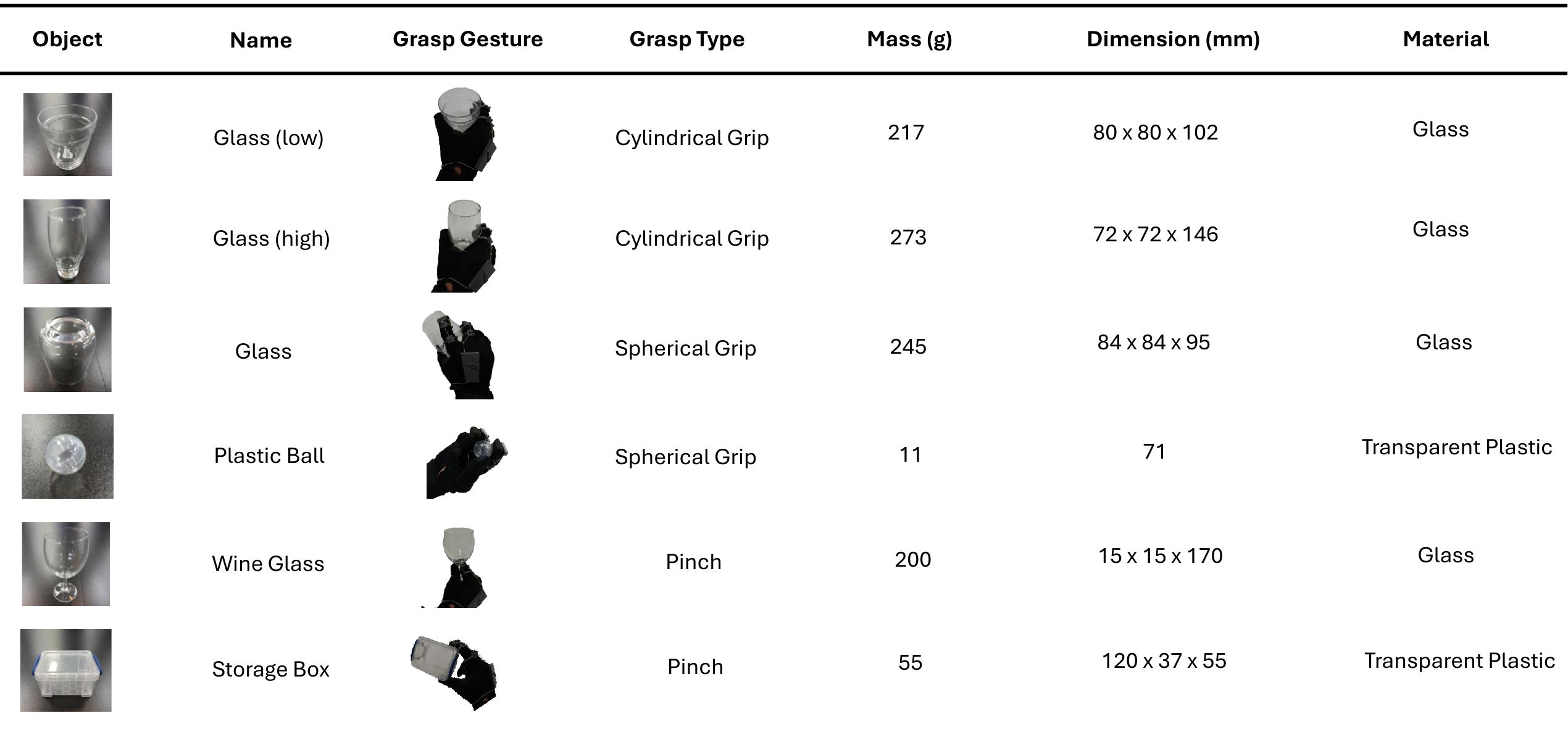}
\caption{The dataset comprises six transparent objects, categorized into three groups, with users performing grasps using three distinct grip types, including Cylindrical Grip, Spherical Grip, and Pinch. The mass, dimensions, and material composition of the objects can significantly impact grasping performance.}
\label{fig:dataset}
\vspace{-0.35cm}
\end{figure*}

\subsection{Dataset}
\vspace{-0.1cm}
Six objects used in the study were chosen to represent three distinct grasp types: cylindrical grip, spherical grasp, and pinch (Fig. \ref{fig:dataset}). These classifications are based on the object set, as described by Jiang et al. \cite{jiang2023robotic}, to ensure heterogeneity across attributes such as shape, mass, dimensions, and material composition. Participants were tasked with grasping the glass (low) and glass (high) for the cylindrical grip. The spherical grip was used for the small plastic ball and the glass in an upside-down orientation. Finally, the pinch was assigned to the wine glass and small storage box. This selection allows for a comprehensive evaluation of grasping performance across varying object characteristics.

\subsection{Participants}
To evaluate the grasping performance of the designed multimodal exoglove, six healthy, right-handed participants were recruited, including four males and two females, with an average age of 25.0±6 years, weight of 75±14 kg, and height of 1.80±0.22 m. All participants demonstrated normal hand motor function. They were instructed to perform grasping tasks while ensuring that these actions caused no discomfort or pain. 

The study received ethical approval (\textit{MRPP-23/24-40750}) from the University and the College Research Ethics Committee (CREC) at King’s College London. Participants were thoroughly briefed on the study's objectives and procedures, and all provided written informed consent prior to participation.

\subsection{Research Procedure}
\vspace{-0.2cm}
\begin{figure}[t]
\centering
\includegraphics[width=\textwidth]{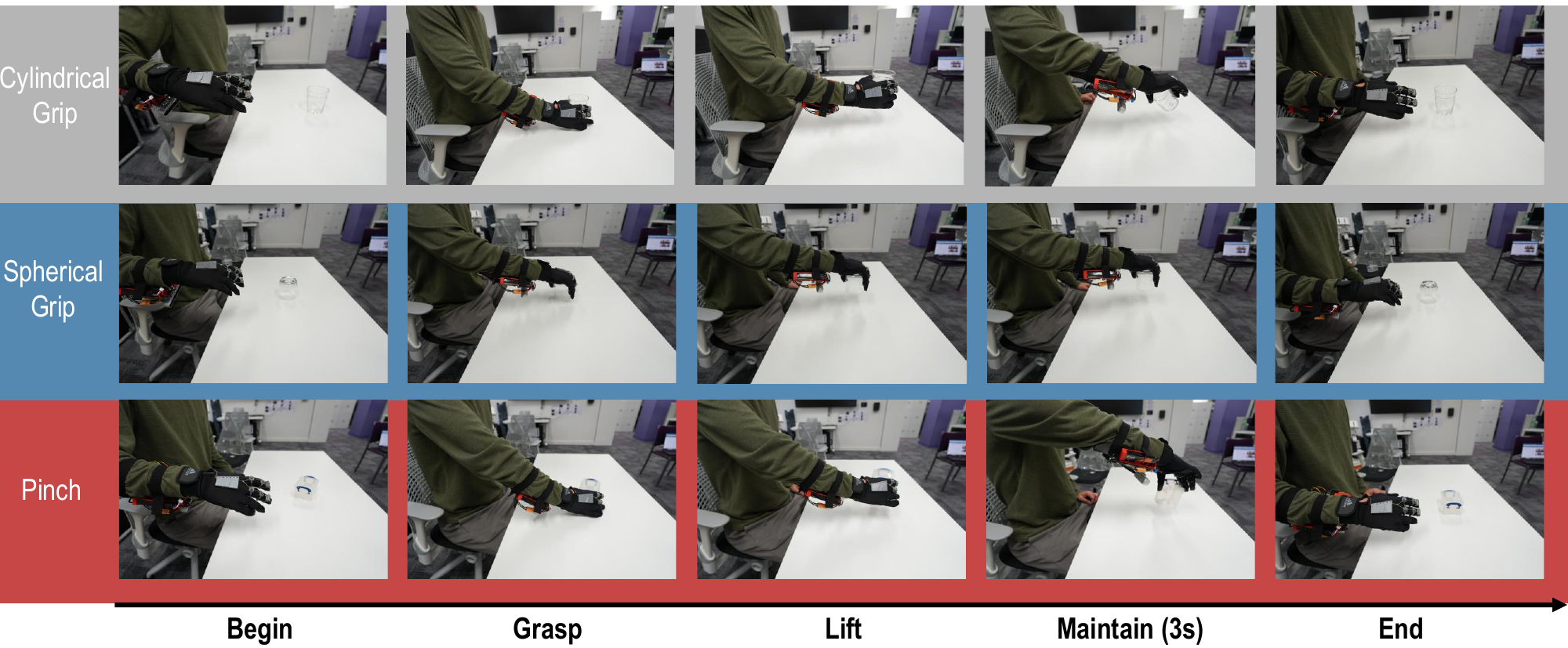}
\caption{\textbf{Five stages of the research procedure:} (1) \textbf{Begin:} The user starts seated at a table, gives the "grip" command, and moves their hand toward the object. (2) \textbf{Grasp:} When the distance between the camera and the object is less than 400mm for 2 seconds, the system triggers the grasp. (3) \textbf{Lift:} Once the motor stops and the grasp stabilizes, the user lifts the object. (4) \textbf{Maintain:} For the Cylindrical and Spherical grips, the user rotates their wrist to a palm-down position and holds this for three seconds. (5) \textbf{End:} The user places the object back on the table, gives the "release" command to reverse the motor, and then issues the "stop" command to halt the motor.} 
\label{fig:interact}
\vspace{-0.6cm}
\end{figure}

During the protocol, participants were seated next to a table (Fig. \ref{fig:overview}). They were instructed to keep their hands relaxed without applying any force on the object during the grasp. A researcher handed the objects to the participants, who were then asked to hold the grasp for three seconds (Fig. \ref{fig:interact}). Afterwards, they rotated their hand 90° to a palm-down position, maintaining the grasp for an additional three seconds before the system initiated the release. Additionally, the researcher recorded how often the camera made contact with the desk to evaluate whether placing the sensor underneath caused any discomfort to users.

\begin{figure}[t]
\centering
\includegraphics[width=\textwidth]{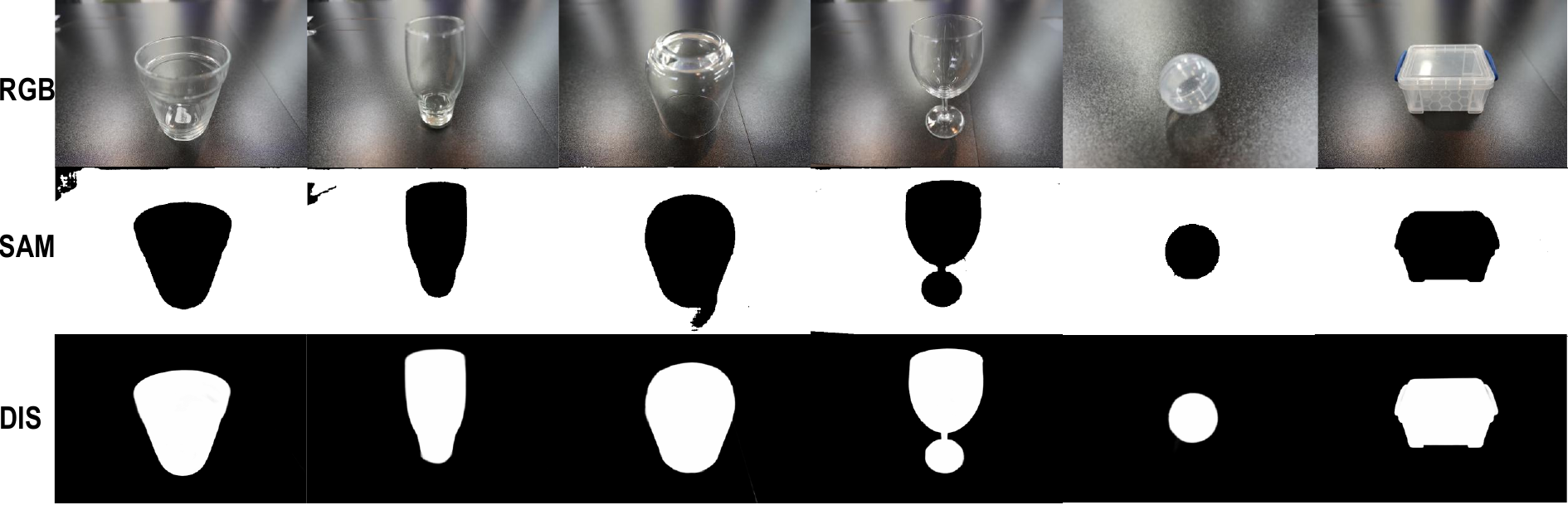}
\caption{The first row displays RGB images of six transparent objects from the dataset. The second row shows masks generated by Segment Anything (SAM), where SAM demonstrates sensitivity to image edges (e.g., Glass (high), Glass (low), and Wine Glass) and object contours, leading to inaccurate segmentation. In contrast, the third row illustrates the masks produced by DIS-Net, which focuses specifically on the outer contours of the transparent objects.}
\label{fig:dis_result}
\vspace{-0.6cm}
\end{figure}

Following the protocol proposed by \cite{maldonado2023fabric}, each object was grasped three times in each control mode to assess the multimodal exoglove’s ability to perform the grasping task accurately (Grasping) and its capability to hold the object (Maintaining). The Grasping and Maintaining scores were evaluated from the moment the object was grasped until it was released. The Grasping score was rated on a scale of 0, 0.5, and 1, where 0 indicated a failure to grasp the object, 0.5 indicated an incorrect grasp type but a successful grip, and 1 indicated a correct grasp. The Maintaining score was similarly rated from 0 to 1, with 0 assigned if the object was dropped, 0.5 if movement occurred, and 1 if the object remained stable. Scores for each object within a grip type were combined to derive a final score for each grasp. A cumulative GAS \cite{llop2019anthropomorphic} was then calculated, representing the overall performance of the multimodal framework in executing grasps. This score was expressed as a percentage, with a GAS of 100\% representing performance equivalent to a healthy individual not wearing a hand exoskeleton.

\subsection{Transparent Object Segmentation}

Fig. \ref{fig:dis_result} visualizes the segmentation outputs for six transparent objects using DIS-Net with the pre-trained weight. Despite not being specifically fine-tuned on transparent objects, DIS-Net accurately captures their contours. It achieves zero-shot segmentation across a range of transparent objects with diverse shapes, laying the groundwork for grasping tasks in open-world environments. In addition, the Segment Anything Model (ViT-B SAM) \cite{kirillov2023segany} was applied to segment the same six objects in the dataset. The results indicate that SAM struggles with contour inference for transparent objects (e.g., an upside-down glass) and certain image frames (e.g., a wine glass). Furthermore, SAM’s model size is more than twice that of DIS-Net, making it less suitable for downstream grasping tasks when evaluated on both accuracy and performance.
\begin{figure}[t]
\centering
\includegraphics[width=\textwidth]{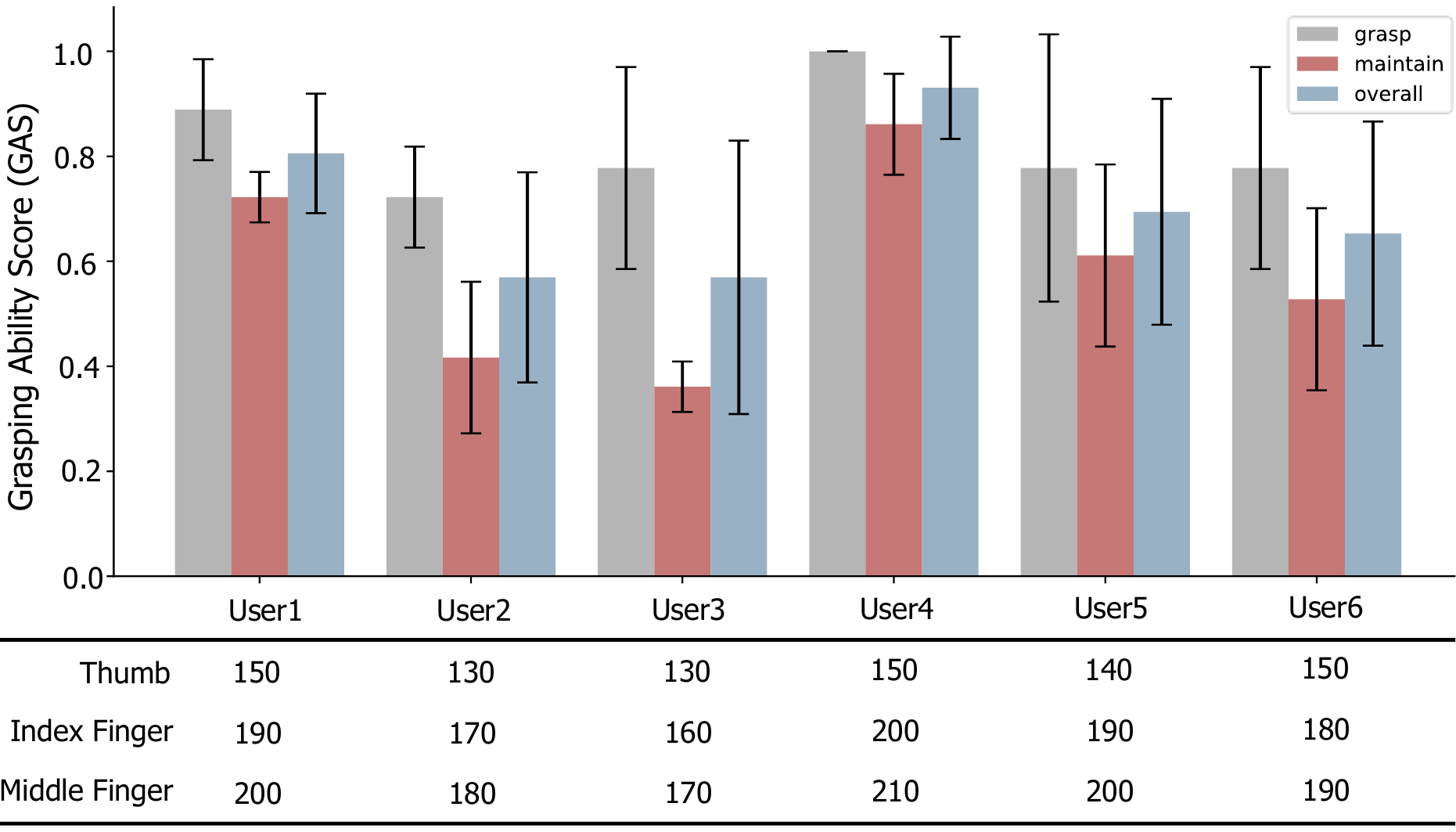}
\caption{The average GAS for six users performing three different grip types on six transparent objects. Grey, red, and blue bars represent the Grasping, Maintaining, and overall GAS, respectively. The table below provides the measured distances (mm) from each user’s palm to the tips of the thumb, index finger, and middle finger.}
\label{fig:gas_result}
\vspace{-0.4cm}
\end{figure}

\subsection{Grasping Ability Score}
The GAS for six participants using the multimodal soft exoglove to perform three types of grasps on six transparent objects (Fig. \ref{fig:gas_result}). On average, the proposed framework achieved a grasping score of $80.4 \pm 5.4\%$, a maintaining score of $60.41 \pm 6.18\%$, and an overall GAS of $70.37 \pm 3.96\%$. Compared to the exoglove presented in \cite{yurkewich2022integrating}, which achieved a GAS score of $53\%$, the proposed system demonstrates a $17.37 \pm 3.96\%$ improvement. While \cite{yurkewich2022integrating} included participants with mild hand impairments, in the study, users did not apply any force during the grasping tasks. Similarly, when compared to the exoglove designed in \cite{maldonado2023fabric}, which achieved a GAS of $80.18 \pm 0.23\%$, the performance is lower by $9.18\%$. However, it is important to note that both \cite{yurkewich2022integrating} and \cite{maldonado2023fabric} evaluated grasping non-transparent objects. Given the material characteristics of transparent objects, such as lower surface friction and distinct weight distribution, grasping them poses additional challenges. 

The distance (mm) was measured from the wrist to the tips of the thumb, index finger, and middle finger. The soft exoglove was built using a 2XL-sized glove, with 3D-printed units attached at the finger joints to route wires. Consequently, the fit between the user’s hand size and the glove significantly impacted grasping success rates. Participants with larger hands, such as User1 and User4, achieved higher GAS scores (Fig. \ref{fig:gas_result}). This issue was particularly evident in the maintaining task, where participants rotated their wrists to a palm-down position. The low friction coefficient of the transparent object surfaces led to frequent slippage or object drops. Moreover, the success rate for grasping transparent plastic objects was higher than that for glass objects.

\begin{table}[t]
    \centering
    \renewcommand{\arraystretch}{1.5}
    \caption{Comparison of the grasping, maintaining, and overall GAS scores for three grasp types using the multimodal soft exoglove.}
    \label{tab:gas_griptype}
    \resizebox{\columnwidth}{!}{
    \begin{tabular}{lccc}
        \hline
        \textbf{Grasp type} & \textbf{Grasping score (\%)} & \textbf{Maintaining score (\%)} & \textbf{GAS score (\%)} \\ \hline
        Cylindrical grip   & 94.44 $\pm$ 3.93   & 68.06 $\pm$ 5.8  & 81.25 $\pm$ 3.37 \\ \hline
        Spherical grip   & 61.11 $\pm$ 7.46   & 48.61 $\pm$ 7.32  & 54.86 $\pm$ 4.72 \\ \hline
        Pinch   & 91.66 $\pm$ 4.81   & 58.33$\pm$ 5.42   & 75.00 $\pm$ 3.78  \\ \hline
        Average   & 80.4 $\pm$ 5.4   & 60.41$\pm$ 6.18   & 70.37 $\pm$ 3.96  \\ \hline
    \end{tabular}
    }
\vspace{-0.4cm}
\end{table}



For the spherical grip, where the palm is consistently positioned downward during both grasping and maintaining tasks, the challenge for the exoglove in grasping transparent objects is heightened. As shown in Table \ref{tab:gas_griptype}, the GAS score for the spherical grip was $54.86 \pm 4.72\%$, significantly lower than that of the cylindrical grip ($81.25 \pm 3.37\%$) by approximately $26\%$, and lower than the pinch grip ($75.00 \pm 3.78\%$) by around $20\%$. The primary reason for this lower performance is that in the spherical grip, the palm faces downward, causing only the fingertips to provide friction. In contrast, other grip types such as the cylindrical grip allow the palm to partially contribute friction, enhancing stability. Repeated trials also exhibited relatively high variability due to this limitation.

\section{Discussion}
\vspace{-0.35cm}
This study introduces a multimodal soft exoglove designed to assist individuals with individuals with neurodegenerative diseases in grasping transparent objects. To validate the glove's ability, GAS tests was conducted, during which six healthy participants participated. Participants were required to keep their hands relaxed and unable to apply any force on the object during the grasp. 

The experiments show that in assistive grasping, even with missing depth information in the interior regions of transparent objects, grasp points can still be computed by leveraging normal discontinuities and thickness variations at the object’s edges, effectively triggering PID control. However, future work will further investigate whether depth completion can further improve assistive grasping accuracy by integrating depth completion algorithms into the \emph{MultiClear} framework.

This study evaluated only three grasp types. Future research will assess the glove's effectiveness across a broader range of grasp types, such as the hook grip. Furthermore, given that some transparent objects in daily life are deformable, upgrading to a more powerful motor and developing an optimized PID control algorithm for deformable object grasping will be key priorities for future research.

Future work will also focus on integrating the multimodal exoglove with digital tools for individuals with neurodegenerative disorders towards self-management of their condition\cite{huo2024co}. Based on user feedback, the glove's design and algorithms will be further tested to better support individuals with neurodegenerative disorders in completing daily tasks.



\section{Conclusion}

As the importance of exogloves for assisting individuals with neurodegenarative disorders in completing daily activities tasks continues to grow, the design and development of soft exogloves with intuitive, user-friendly interaction to effectively support transparent object grasping remains an open challenge. This paper presents a multimodal framework \emph{MultiClear} designed to improve the efficiency of grasping transparent objects by combining RGB, depth, and auditory modalities. The framework is embedded into a soft exoglove that bridges the gap between high-level contextual awareness and low-level PID control. Experimental results show that the proposed exoglove achieves a GAS of 70.37±3.96\%, illustrating the possible effectiveness of zero-shot segmentation for transparent objects using a vision foundation model based on dichotomous segmentation. Future work aims to explore whether integrating depth completion into the \emph{MultiClear} framework can further enhance the accuracy of assistive grasping in individuals with motor disorders, testing the system's effectiveness on post-stroke survivors for instance, refining both the design and functionality to better support daily tasks.

%
%
\bibliographystyle{splncs04}
\bibliography{references}

\end{document}